\documentclass[11pt]{article}

\usepackage[final]{acl}

\usepackage{iftex}
\ifPDFTeX
  \usepackage{times}
  \usepackage[T1]{fontenc}
  \usepackage[utf8]{inputenc}
\else
  \usepackage{fontspec}
  \IfFontExistsTF{Noto Serif Bengali}
  {\newfontfamily\bnf[Script=Bengali]{Noto Serif Bengali}}
  {\newfontfamily\bnf[Script=Bengali]{FreeSerif.otf}}
\fi

\usepackage{times}
\usepackage{latexsym}

\usepackage[T1]{fontenc}

\usepackage[utf8]{inputenc}
\usepackage{xspace}

\usepackage{microtype}

\usepackage{inconsolata}

\usepackage{graphicx}
\usepackage{graphicx}
\usepackage{amsmath}
\usepackage{booktabs}
\usepackage{comment}
\usepackage{subcaption}
\usepackage{tikz}
\usetikzlibrary{arrows.meta,shapes.geometric,calc,fit,backgrounds}

\usepackage{xcolor,colortbl}
\usepackage{tikz}
\usepackage{tcolorbox}\tcbuselibrary{breakable,skins}
\usepackage{tabularx,array,amssymb,pifont,needspace}
\usepackage{enumitem}

\usepackage{inconsolata}
\usepackage{xcolor}
\usepackage{booktabs}
\usepackage{tikz}
\usetikzlibrary{positioning, arrows.meta}

\newcommand{\name}{BanglaPLAN\xspace}
\newcommand{\rot}[1]{\rotatebox{85}{#1}} 

\newtcolorbox{takeaway}{colback=black!4,colframe=black!55,boxrule=0.5pt,
    arc=2.5mm,left=5pt,right=5pt,top=3pt,bottom=3pt}

\definecolor{promptgreen}{RGB}{112,173,71}
\definecolor{phgray}{RGB}{90,90,90}
\definecolor{stepblue}{RGB}{68,114,196}
\newcommand{\ph}[1]{\textcolor{phgray}{\{#1\}}}

\newcommand{\masudok}[1]{%
  \textcolor{teal}{\textbf{MASUDOK:}} \textcolor{orange}{#1}}
\newcommand{\masud}[1]{%
  \textcolor{teal}{MASUD:} \textcolor{violet}{#1}}
\newcommand{\jami}[1]{%
  \textcolor{blue}{JAMI: #1}}
\newcommand{\mimon}[1]{%
  \textcolor{red}{MIMON:} \textcolor{purple}{#1}}

\newcommand{\eatreminders}{%
  \renewcommand{\masud}[1]{}%
  \renewcommand{\jami}[1]{}%
  \renewcommand{\mimon}[1]{}%
  \renewcommand{\masudok}[1]{##1}%
}

\eatreminders

\title{\textsc{\name}: A Benchmark for Constraint-Based Planning and Decision-Making in Bangla}

\author{
    \textbf{Md Rayhanul Masud}, 
    \textbf{S M Maksudul Alam}, 
    \textbf{Mohammad Jaminur Islam} \\ 
    University of California, Riverside \\
}

\begin{document}
\maketitle

\begin{abstract}
\jami{This is a great abstract: My concern is as follows 
1. Does it sound like too verbose ?
2. Is it somehow hiding or blurring the motivation?
Motivation: LLM models' performance does not translate to cross-lingual cases; we analyse the gap for the Bangla language with a benchmark dataset created by us, or something similar. 
Note: I feel like we have too generic statements at the start.
}

Large language models (LLMs) increasingly help users arrange appointments, plan journeys, and allocate limited resources. 
These tasks require models to satisfy constraints on time, cost, capacity, and order, or determine that no feasible solution exists. 
Bangla is the world's seventh most widely spoken language, and existing benchmarks cover many language and reasoning tasks.
However, benchmarks for evaluating planning and decision-making in Bangla under explicit constraints remain lacking.
We introduce \textsc{\name}, a benchmark of 1,000 problems across ten task families grounded in everyday life and institutions in Bangladesh. 
It includes feasible and infeasible problems in four language registers: everyday, formal, colloquial, and Bangla--English mixed. 
Answers cover eight structured types, including schedules, routes, allocations, and ordered steps, in a common JSON format.
We construct the benchmark through language-model-assisted drafting and independent human validation. 
We evaluate 11 open-weight and proprietary model configurations using exact match, separately reporting feasible-case accuracy, infeasibility detection, and format errors.
We find a large performance gap: GPT-6.1 Sol achieves 98.7\% accuracy and Qwen3.5-27B with thinking reaches 80.3\%, while direct-answer configurations score at most 30.9\%. 
Seven of eleven configurations perform below the always-``infeasible'' baseline.
Our findings also suggest that harder problems primarily reduce plan-construction accuracy, while infeasibility detection remains more robust.
\textsc{\name} provides a reproducible basis for evaluating planning and decision-making in Bangla. \textsc{\name} is available at: https://github.com/mrayhanulmasud/banglaplan.
\end{abstract}

\section{Introduction}
\label{sec:introduction}
\jami{
Here, I feel like the first paragraph is again describing generic statements, and I feel like this is somehow diluting the core scope or motivation for a Bangla benchmark. I could be wrong, Should we invest more on why Bangla Benchmark or language specific benchmark is needed then invest if its solving anything specific regarding the scoped motivation then showing our contribution etc. Feel free to disagree.
}
Large language models (LLMs) can follow natural-language instructions and perform multi-step reasoning \cite{brown2020language,wei2022chain}.
Their evaluation has therefore expanded from question answering and text generation to planning and constrained decision making \cite{valmeekam2023planbench,zheng2024naturalplan,tso2026constraintbench,zhao2026planningbench,zhang2026deepplanning}, including scheduling \cite{zheng2024naturalplan,tso2026constraintbench}, travel planning \cite{zheng2024naturalplan,xie2024travelplanner}, and resource allocation \cite{tso2026constraintbench}.
A valid response must satisfy every constraint \cite{xie2024travelplanner,tso2026constraintbench} or, when that is impossible, recognize that no feasible solution exists \cite{qi2026trek}.

Existing planning benchmarks span formal, natural-language, and real-world settings \cite{valmeekam2023planbench,zheng2024naturalplan,xie2024travelplanner}.
Recent work covers practical tasks \cite{zhao2026planningbench}, constrained optimization \cite{tso2026constraintbench}, long-horizon planning \cite{zhang2026deepplanning}, and infeasible problems \cite{qi2026trek}.
Bangla is the seventh most widely spoken language, yet it is rarely represented in planning benchmarks.
Existing Bangla benchmarks cover subject knowledge \cite{joy2025bnmmlu}, question answering \cite{shafayat2024benqa}, mathematical reasoning \cite{prama2025banglamath}, and social interaction \cite{sijan2026banglasocialbench}.
Multilingual planning and workflow benchmarks do not include Bangla \cite{jung2025multiplan,ranaldi2026worldbench}, while other multilingual evaluations include Bengali but target different capabilities \cite{pahuja2026gaiamaps,jain2025voiceagentbench}.
These benchmarks leave a gap in evaluating models' ability to plan and make decisions under explicit constraints in Bangla.

\begin{figure*}[t]
  \centering
  \includegraphics[width=0.9\textwidth]{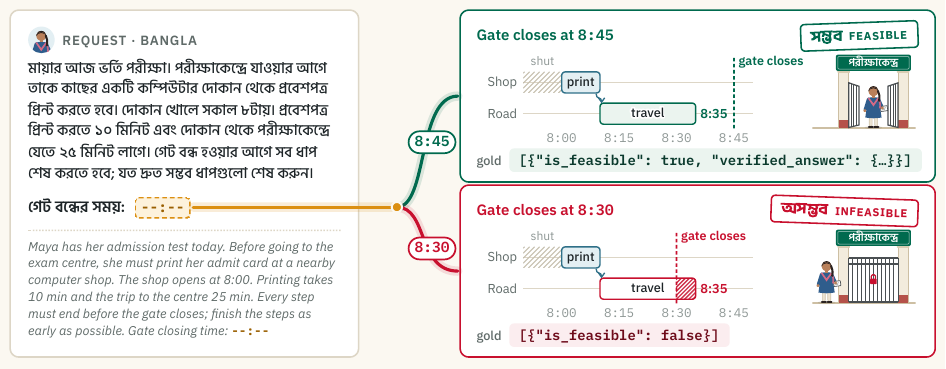}
  \caption{A \textsc{\name} problem and its near-boundary infeasible variant. Left: the Bangla request with an English translation. A student needs 10 minutes to print her admit card at a shop that opens at 8:00, followed by 25 minutes of travel to the exam centre. Right: a gate-closing time of 8:45 permits the gold plan; changing it to 8:30 makes the request infeasible, with gold answer \texttt{is\_feasible: false}.}
  \label{fig:teaser}
\end{figure*}

\paragraph{Research questions.}
This gap motivates two research questions:
\begin{enumerate}
    \item \textbf{RQ1: Overall performance.} How well do current models construct valid plans and identify infeasible requests under explicit constraints in Bangla?
    \item \textbf{RQ2: Variation and errors.} How does performance vary across benchmark dimensions, and which failure modes dominate?
\end{enumerate}

\paragraph{Contributions.} We make the following contributions to address these questions:
\begin{itemize}[leftmargin=15pt, nosep]
    \item \textbf{A Bangla planning benchmark.} We introduce \textsc{\name}, a benchmark of 1,000 problems authored directly in Bangla across ten task families grounded in Bangladesh. It contains 744 feasible and 256 infeasible problems, with 300 easy, 300 medium, and 400 hard instances. The problems span four language registers and eight structured answer types in a common JSON format. Section~\ref{sec:benchmark} describes the construction process. Figure~\ref{fig:teaser} presents an example.
    \item \textbf{Validated reference answers.} We combine language-model-assisted drafting with independent human validation to establish canonical answers for deterministic exact-match scoring.
    \item \textbf{Model evaluation and error analysis.} We evaluate 11 open-weight and proprietary model configurations, separating plan accuracy, infeasibility detection, and format errors, and analyzing performance across benchmark dimensions.
\end{itemize}



\jami{Great summarization of the result. Should we use forward reference for detail result. For example, if reviewer wonder format error what does it look like they can go to specific section and find the details.}

\paragraph{Results.} Our evaluation yields the following key findings:
\begin{itemize}[leftmargin=15pt, nosep]
    \item \textbf{RQ1. Test-time reasoning yields larger gains than scaling.} Models using test-time reasoning perform best: GPT-6.1 Sol achieves 98.7\% exact match, GPT-6 Luna 85.1\%, and Qwen3.5-27B with thinking 80.3\%. Direct-answer configurations reach at most 30.9\%, and seven of eleven configurations score below the 25.6\% always-``infeasible'' baseline. Enabling thinking with the same 27B weights adds 49.4 percentage points, exceeding the gain from scaling Qwen3.5 from 2B to 27B.\jami{To me: What I want to see is a key result and takeaway; here, so many numbers might be a little heavy to connect. Besides, we are actually presenting the same result to the result section, so we could present the key meaning or finding instead to many numbers.}\masud{fixed. I agree with you. However, I think we should report numbers that provide a landscape of how results look like. To make it obvious, we added keytakeaway line at the beginning.}
    \item \textbf{RQ2. Difficulty affects planning more than infeasibility detection.} Harder problems \jami{Should we refer to the problem type in our benchmark dataset- as I feel what kind of problems we are referring as harder and who categorizes them as harder}\masud{we defined later in Section 3. However, we added modification in contribution.} reduce plan accuracy more than infeasibility detection: GPT-6 Luna's plan accuracy falls from 96.0\% to 69.0\%. Constructed answers, such as schedules, are harder than label selection. Format errors further limit performance, affecting 70.3\% of Gemma~3 outputs.
    \item \textbf{Translation and fine-tuning offer limited gains in plan construction.} English prompts and gold-answer fine-tuning mainly reduce format errors and change the frequency of ``infeasible'' responses, while improving plan construction by only a few percentage points. \mimon{not sure if we should merge contribution and result segments.}
\end{itemize}

\section{Related Work}
\label{sec:related}

\begin{table*}[t]
\centering
\small
\setlength{\tabcolsep}{3pt}
\caption{The ten task families in \textsc{\name}, each comprising ten problem types with ten instances per type.\jami{Table caption should be at top?}\masud{fixed}}
\begin{tabular}{@{}l >{\raggedright\arraybackslash}p{0.30\textwidth} @{\hspace{10pt}} l >{\raggedright\arraybackslash}p{0.30\textwidth}@{}}
\toprule
\textbf{Family} & \textbf{Representative problem types} & \textbf{Family} & \textbf{Representative problem types} \\
\midrule
F01 Eligibility   & scholarships, exams, leave, mobile-money rules & F06 Sequencing  & recipes, construction, crop calendars \\
F02 Procedures    & NID correction, passports, land mutation        & F07 Allocation  & beds, relief rice, irrigation, Qurbani meat \\
F03 Appointments  & clinics, biometrics, banks, legal aid           & F08 Procurement & haat shopping, Iftar supplies, tenders \\
F04 Journeys      & Eid travel, ferries, hospital referrals         & F09 Logistics   & ports, ferry berths, waste collection \\
F05 Rosters       & exams, invigilation, garment shifts             & F10 Emergency   & evacuation, triage, ambulance dispatch \\
\bottomrule
\end{tabular}
\label{tab:families}
\end{table*}

\paragraph{Planning benchmarks.}
LLM planning benchmarks, mainly in English, vary in problem formality and task coverage. \textsc{PlanBench} evaluates plan generation and reasoning through PDDL problems \citep{valmeekam2023planbench}. \textsc{Natural Plan} covers natural-language trip, meeting, and calendar planning \citep{zheng2024naturalplan}, while \textsc{TravelPlanner} evaluates travel plans against real-world constraints \citep{xie2024travelplanner}. \textsc{PlanningBench} \citep{zhao2026planningbench} broadens coverage to six problem families, and \textsc{DeepPlanning} examines long-horizon travel and shopping tasks \citep{zhang2026deepplanning}. Translating English datasets can introduce errors and lose cultural nuance \citep{salazar2026kaleidoscope,romanou2025include}, motivating benchmarks rooted in local contexts. \textsc{\name} addresses this need with problems authored directly in Bangla and grounded in everyday life in Bangladesh.

\paragraph{Multilingual planning.}
Multilingual planning evaluation remains narrow.
\textsc{Multi-Plan} compares Korean and English travel planning \citep{jung2025multiplan}, \textsc{ChinaTravel} targets Chinese \citep{shao2026chinatravel}, and \textsc{Ticket-Bench} covers six European languages, with 15\% of its queries infeasible \citep{almeida2025ticketbench}.
\textsc{WorldBench} builds scheduling, travel, and expense tasks natively in seven languages and eight cultures, but includes no South Asian language \citep{ranaldi2026worldbench}.
Two resources include Bangla, yet neither targets locally grounded planning under explicit constraints:
\textsc{GAIA-MAPS} diagnoses multi-agent planning failures across eleven languages, with Bangla tasks produced by machine translation of GAIA \citep{pahuja2026gaiamaps},
and \textsc{VoiceAgentBench} tests spoken tool calls built from translated Indian-context queries in seven languages \citep{jain2025voiceagentbench}.

\paragraph{Bangla evaluation.}
Existing Bangla benchmarks cover subject knowledge \citep{joy2025bnmmlu}, science question answering and reasoning in Bangla and English \citep{shafayat2024benqa}, reading comprehension \citep{rony2024banglaquad}, school mathematics \citep{prama2025banglamath}, and sociopragmatic and cultural alignment \citep{sijan2026banglasocialbench}.
Prior work also studies whether script fragmentation and prompt format explain the English--Bangla performance gap \citep{bhowmik2025script}.
A few Bangla datasets contain items with no valid answer: \textsc{NCTB-QA} labels 42.75\% of its questions unanswerable \citep{eyasir2026nctbqa}, and \textsc{GSM-Plus-BN} includes math problems with missing information \citep{paul2026gsmplusbn}.
Both, however, test reading or arithmetic rather than planning.
To our knowledge, no Bangla benchmark evaluates planning and decision making under explicit constraints.

\section{The \textsc{\name} Benchmark}
\label{sec:benchmark}

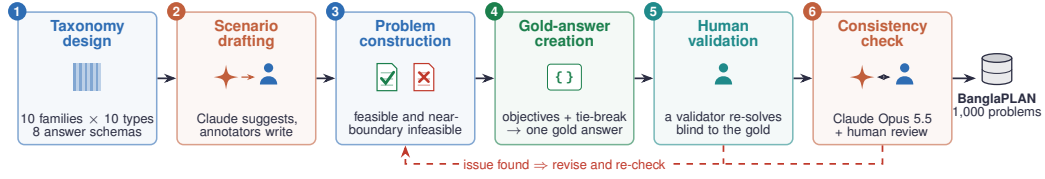
\begin{figure*}[t]
\centering
\resizebox{0.9\textwidth}{!}{
\definecolor{pAnn}{HTML}{2E6DB4}   
\definecolor{pLLM}{HTML}{C15F3C}   
\definecolor{pVal}{HTML}{1F8A8A}   
\definecolor{pGold}{HTML}{1B7F4C}  
\definecolor{pInf}{HTML}{C0392B}   
\definecolor{pInk}{HTML}{2B2D42}
\tikzset{
  stage/.style={draw=#1!60,fill=#1!5,rounded corners=3pt,line width=0.7pt,minimum width=2.25cm,minimum height=2.15cm},
  stitle/.style={font=\sffamily\bfseries\scriptsize,text=#1,align=center,text width=2.15cm,inner sep=1pt},
  scap/.style={font=\sffamily\tiny,text=pInk,align=center,text width=2.2cm,inner sep=1pt},
  num/.style={circle,fill=#1,text=white,font=\sffamily\bfseries\tiny,inner sep=1.1pt},
  flow/.style={-{Stealth[length=2mm]},line width=0.8pt,pInk},
  person/.pic={\fill[#1] (0,0.15) circle (0.075);
               \fill[#1] (-0.14,-0.12) .. controls (-0.14,0.06) and (0.14,0.06) .. (0.14,-0.12) -- cycle;},
  spark/.pic={\fill[#1] (0,0.2) .. controls (0.03,0.03) .. (0.2,0) .. controls (0.03,-0.03) .. (0,-0.2)
                       .. controls (-0.03,-0.03) .. (-0.2,0) .. controls (-0.03,0.03) .. (0,0.2) -- cycle;},
  doc/.pic={\draw[fill=white,draw=#1,line width=0.55pt] (-0.16,-0.22) -- (0.16,-0.22) -- (0.16,0.14) -- (0.08,0.22) -- (-0.16,0.22) -- cycle;
            \foreach \y in {0.07,-0.03,-0.13} \draw[#1!55,line width=0.45pt] (-0.1,\y) -- (0.1,\y);},
  grid/.pic={\foreach \i in {0,...,9}{\foreach \j in {0,...,9}{
               \pgfmathsetmacro\c{mod(\i,2)==0?60:32}
               \fill[#1!\c] (-0.22+\i*0.045,0.22-\j*0.045) rectangle ++(0.037,-0.037);}}},
  json/.pic={\node[draw=#1,fill=white,rounded corners=1.5pt,font=\ttfamily\bfseries\scriptsize,text=#1,minimum width=0.55cm,minimum height=0.42cm,inner sep=1pt] {\{\,\}};},
  db/.pic={\draw[fill=#1!10,draw=#1,line width=0.6pt] (-0.24,0.18) -- (-0.24,-0.18) arc (180:360:0.24 and 0.07) -- (0.24,0.18);
           \draw[fill=#1!22,draw=#1,line width=0.6pt] (0,0.18) ellipse (0.24 and 0.07);
           \draw[#1,line width=0.45pt] (-0.24,0.0) arc (180:360:0.24 and 0.07);},
  okmark/.pic={\draw[pGold,line width=1.3pt,line cap=round] (-0.09,0) -- (-0.02,-0.08) -- (0.11,0.1);},
  nomark/.pic={\draw[pInf,line width=1.2pt,line cap=round] (-0.075,-0.075) -- (0.075,0.075) (-0.075,0.075) -- (0.075,-0.075);},
}
\begin{tikzpicture}[x=1cm,y=1cm]
\foreach \i/\c/\t/\cp [count=\n] in {
   0/pAnn/{Taxonomy\\[-1pt]design}/{10 families $\times$ 10 types\\[-1pt]8 answer schemas},
   1/pLLM/{Scenario\\[-1pt]drafting}/{Claude suggests,\\[-1pt]annotators write},
   2/pAnn/{Problem\\[-1pt]construction}/{feasible and near-\\[-1pt]boundary infeasible},
   3/pGold/{Gold-answer\\[-1pt]creation}/{objectives + tie-break\\[-1pt]$\rightarrow$ one gold answer},
   4/pVal/{Human\\[-1pt]validation}/{a validator re-solves\\[-1pt]blind to the gold},
   5/pLLM/{Consistency\\[-1pt]check}/{Claude Opus 5.5\\[-1pt]+ human review}}{
  \node[stage=\c] (s\n) at (\i*2.58,0) {};
  \node[num=\c] at (s\n.north west) {\n};
  \node[stitle=\c,anchor=north] at ($(s\n.north)+(0,-0.1)$) {\t};
  \node[scap,anchor=south] at ($(s\n.south)+(0,0.1)$) {\cp};
}
\def\yI{0.02}
\pic at ($(s1)+(0,\yI)$) {grid=pAnn};
\pic at ($(s2)+(-0.33,\yI)$) {spark=pLLM};
\draw[-{Stealth[length=1.3mm]},pLLM,line width=0.6pt] ($(s2)+(-0.08,\yI)$) -- ($(s2)+(0.14,\yI)$);
\pic at ($(s2)+(0.36,\yI)$) {person=pAnn};
\pic at ($(s3)+(-0.3,\yI)$) {doc=pGold}; \pic at ($(s3)+(-0.3,\yI-0.02)$) {okmark};
\pic at ($(s3)+(0.3,\yI)$) {doc=pInf};   \pic at ($(s3)+(0.3,\yI)$) {nomark};
\pic at ($(s4)+(0,\yI)$) {json=pGold};
\pic at ($(s5)+(0,\yI)$) {person=pVal};
\pic at ($(s6)+(-0.33,\yI)$) {spark=pLLM};
\draw[{Stealth[length=1.3mm]}-{Stealth[length=1.3mm]},pInk,line width=0.6pt] ($(s6)+(-0.09,\yI)$) -- ($(s6)+(0.13,\yI)$);
\pic at ($(s6)+(0.36,\yI)$) {person=pAnn};
\foreach \a/\b in {1/2,2/3,3/4,4/5,5/6}{\draw[flow] (s\a.east) -- (s\b.west);}
\draw[flow] (s6.east) -- ++(0.38,0);
\pic at ($(s6.east)+(0.72,0.15)$) {db=pInk};
\node[font=\sffamily\bfseries\tiny,text=pInk,align=center,anchor=north] at ($(s6.east)+(0.72,-0.12)$) {\name\\[-1pt]\mdseries 1,000 problems};
\draw[-{Stealth[length=2mm]},pInf,dashed,line width=0.8pt]
  (s6.south) -- ++(0,-0.3) -| (s3.south);
\draw[pInf,dashed,line width=0.8pt] (s5.south) -- ++(0,-0.3);
\node[font=\sffamily\tiny,text=pInf,fill=white,inner sep=1.5pt] at ($(s4.south)+(0,-0.3)$) {issue found $\Rightarrow$ revise and re-check};
\end{tikzpicture}}
\caption{Construction of \textsc{\name}. Authors define the taxonomy. Annotators select and rewrite Claude-suggested scenarios in Bangla, creating feasible and near-boundary infeasible problems with unique canonical gold answers. An independent annotator solves each problem without seeing the gold answer, followed by a second check using Claude Opus~5.5. Humans resolve disagreements and revise flagged issues (red).}
\label{fig:pipeline}
\vspace{-10pt}
\end{figure*}

\masudok{\textsc{\name} benchmarks constraint-based planning in Bangla across tasks such as scheduling, routing, and allocation. Each instance requires a valid structured answer or an infeasibility decision. Explicit constraints, objectives, and tie-breaking rules determine a unique canonical answer for deterministic scoring without an LLM judge.}

\jami{Do you think it would be great if we say the instances represent different tasks in this intro paragraph, as the next section starts with task definition that feels why all of a sudden we are defining the task and what task we are referring to?while reading later I realized we have have multiple tasks within each instance therefore the task definition need more clarification?}\masud{Fixed.}

\subsection{Task Definition}
\label{sec:task}

Each instance is a tuple $x=(p,C,\mathbf{f},\tau)$, where $p$ is a Bangla prompt specifying hard constraints $C$, ordered objectives $\mathbf{f}=(f_1,\dots,f_K)$, and a tie-breaking rule $\tau$. A \emph{plan} $y$ is a well-formed answer of the required type, such as a schedule, route, or allocation. Let $S_0=S(x)$ \jami{I believe we did not use $S_0$ later; therefore just saying S(x) should be fine, right?} denote all plans satisfying $C$. For feasible instances, we apply the objectives sequentially:
\[
S_k=\operatorname*{arg\,opt}_{y\in S_{k-1}} f_k(y),
\qquad k=1,\dots,K,
\]
where each objective is minimized or maximized as specified in the prompt. The gold answer is
\[
y^\ast(x)=
\begin{cases}
\textsc{infeasible} & \text{if } S(x)=\emptyset,\\
\tau(S_K) & \text{otherwise,}
\end{cases}
\]
where $\tau$ selects one plan from $S_K$. This defines a lexicographic objective order: each objective is optimized only among plans satisfying all earlier objectives. For example, a scheduling problem may prioritize the earliest completion time, then the earliest start of the first task, and finally task-ID order to resolve remaining ties. Some instances require no optimization ($K=0$), as in eligibility decisions where the constraints determine a single answer. The stated objectives and tie-breaking rules ensure a unique gold answer.

The model must return a one-element JSON array. A feasible response takes the form
\begin{center}\scriptsize
\texttt{[\{"is\_feasible":true,"verified\_answer":\{...\}\}]}
\end{center}
and an infeasible response is
\begin{center}\scriptsize
\texttt{[\{"is\_feasible":false\}]}
\end{center}
The \texttt{verified\_answer} object follows one of eight schemas: classification, ordered steps, schedule, route, allocation, matching, selection, or ranking. A response is correct if it parses as JSON, conforms to the required schema, and matches the gold answer's JSON representation after deterministic canonicalization. Parsing and schema violations are recorded as format errors.

\paragraph{Coverage.}
\textsc{\name} contains 1,000 problems across ten task families, each with ten problem types and ten instances per type (Table~\ref{tab:families}). It includes 744 feasible and 256 infeasible instances, divided into 300 easy, 300 medium, and 400 hard problems. Prompts span four registers: everyday, formal, colloquial, and Bangla--English code-mixed. Answers follow eight types: allocation, classification, matching, ordered steps, ranking, route, schedule, and selection. Appendix~\ref{sec:appendix-breakdowns} reports counts by register and answer type.

\paragraph{Linguistic variation.}
Prompts vary in how they express numbers, dates, and times. 
Numbers appear as Bangla digits, English digits, or words; dates may follow the Bengali calendar; and times may be relative (``next Thursday'') or expressed conversationally (``half past ten''). Relative-date problems specify a reference date and time, while Bengali-calendar problems provide the information needed for conversion. Output values consistently use ASCII digits and ISO-8601 timestamps with the \texttt{+06:00} offset. Each problem has an English version preserving its entities, quantities, constraints, and gold answer as we describe later in Section~\ref{sec:evaluation}. The Bangla version is the original, authored directly in Bangla.

\subsection{Benchmark Construction}
\label{sec:construction}

We construct \textsc{\name} in six stages: taxonomy design, scenario drafting, problem construction, gold-answer creation, human validation, and a model-assisted consistency check. Figure~\ref{fig:pipeline} illustrates the benchmark construction pipeline.

\paragraph{Annotators and pilot.}
Five native Bangla speakers with relevant experience volunteered: three drafted problems and two independently validated them. A 100-problem pilot refined guidelines, answer schemas, and rules for ambiguity, feasibility, and tie-breaking. Appendix~\ref{sec:annotation-guidelines} provides the annotation and validation guidelines.

\paragraph{Stage 1: Taxonomy design.}
We defined ten task families reflecting planning and decision-making needs in Bangladesh, including administrative procedures, river travel, garment shifts, irrigation, and emergency response. Each family contains ten concrete problem types. \mimon{Should we clearly list them here and refer to the examples of each one in the appendix?} Before drafting, we established five constraint categories temporal, spatial, resource-based, logical, and quantitative and the eight answer schemas described in Section~\ref{sec:task}.\jami {Maybe we can break this sentence into two parts, given we are talking about constraints and answers together. Also, we have used and twice, which sounds a little odd.}

\paragraph{Stage 2: Scenario drafting.}
Annotators used Claude Sonnet~5.5 to suggest people, places, resources, rules, and plausible quantities for each problem type. They selected, revised, and checked these suggestions for natural wording, local relevance, and internal consistency. Each scenario was recorded in a structured form specifying entities, identifiers, quantities, time windows, and constraints. Annotators then authored the Bangla prompt directly from this record. Every prompt explicitly states that its names, dates, and rules are fictional.

\paragraph{Stage 3: Problem construction.}
Feasible problems admit at least one plan satisfying all constraints ($S(x)\neq\emptyset$). Annotators created infeasible variants by changing one binding quantity, time window, capacity, or policy rule until no valid plan remained ($S(x)=\emptyset$). We prioritized near-boundary conflicts, such as requiring 187 units when only 186 are available. An instance is \emph{minimally infeasible} only if relaxing its designated conflicting constraint restores feasibility. This design requires precise constraint checking. Some prompts also include irrelevant facts or non-binding preferences to test robustness; these do not affect the gold answer.

\paragraph{Stage 4: Gold-answer creation.}
We establish a unique canonical answer for each feasible instance through three steps. First, we order objectives lexicographically, such as maximizing savings before educational benefit. Second, explicit tie-breaking rules resolve remaining choices. Third, we clarify potentially ambiguous conventions, including whether the start date counts as day~0 and whether $N$ days means $N\times24$ hours.\jami{this is great description of how we got the gold answer: However do we need e.g., (the tie breaking example instance showing the case concretely or referring to the example in the appendix, ambiguous convention: reference to appendix or in-line example)}

We also standardize output representation. Identifiers must match those in the prompt, and zero-valued allocations are omitted. Allocations follow prompt order, while scheduled actions follow start-time order, with ties resolved by the prompt's task-ID order. The scorer parses and canonicalizes valid JSON before exact matching, making scores independent of whitespace and key order.

\subsection{Validation and Difficulty}
\label{sec:validation}

\paragraph{Stage 5: Human validation.}
Each author derived the gold answer using the stated constraints, objectives, and tie-breaking rules. An independent validator then solved the problem from the Bangla prompt alone, without seeing the gold answer, and checked its completeness and clarity. The author and validator resolved disagreements about feasibility, the plan, or the output schema. Revised problems underwent validation again.

\paragraph{Stage 6: Model-assisted consistency check.}
Claude Opus~5.5 independently solved every human-validated problem without access to its gold answer. Mismatches and prompts flagged as ambiguous or incomplete underwent human review. Claude suggested possible causes and revisions, which annotators accepted, edited, or rejected. Revised problems were solved and checked again until humans confirmed that any remaining mismatches were model errors. Claude's outputs served as review signals; human judgment determined the final prompts and gold answers.

\paragraph{Difficulty.}
Annotators recorded the binding constraints and supporting facts required to solve each problem. We define structural difficulty by their total count $k$: easy instances have $k\le 10$, medium instances have $10<k\le 20$, and hard instances have $k>20$. We fixed both thresholds before model evaluation, yielding 300 easy, 300 medium, and 400 hard instances. This measure reflects structural complexity rather than prompt length. 

\section{Evaluation}
\label{sec:evaluation}
\begin{table*}[t]
\centering\small
\setlength{\tabcolsep}{4pt}
\caption{Zero-shot results on 1,000 \textsc{\name} instances (all values in \%). Each feasible or infeasible block sums to 100, up to rounding.
\emph{EM}: exact-match accuracy over all instances.
For feasible instances, \emph{Plan} denotes a correct plan, \emph{Wrong plan} an incorrect plan, and \emph{Abstention} an incorrect ``infeasible'' answer. For infeasible instances, \emph{Flag} denotes a correct ``infeasible'' answer and \emph{False plan} a proposed plan.
\emph{Format} denotes failure to return the required JSON array.
\textbf{Bold} and \underline{underline} mark the best and second-best EM, Plan, and Flag scores.
\textcolor{gray}{Gray} marks EM below the always-``infeasible'' baseline of 25.6.\jami{Table caption should be at top right?}\masud{fixed}}
\begin{tabular}{@{}l rrr | rrrr | rrr@{}}
\toprule
 & \multicolumn{3}{c}{All 1,000 instances} & \multicolumn{4}{c}{744 feasible instances} & \multicolumn{3}{c}{256 infeasible instances} \\
\cmidrule(lr){2-4}\cmidrule(lr){5-8}\cmidrule(l){9-11}
Model & EM$\uparrow$ & Incorrect & Format & Plan$\uparrow$ & \begin{tabular}[b]{@{}c@{}}Wrong\\plan\end{tabular} & \begin{tabular}[b]{@{}c@{}}Absten-\\tion\end{tabular} & Format & Flag$\uparrow$ & \begin{tabular}[b]{@{}c@{}}False\\plan\end{tabular} & Format \\
\midrule
\textit{Always ``infeasible''} & \textit{25.6} & \textit{74.4} & \textit{0.0} & \textit{0.0} & \textit{0.0} & \textit{100.0} & \textit{0.0} & \textit{100.0} & \textit{0.0} & \textit{0.0} \\
\midrule
\multicolumn{11}{@{}l}{\textsc{Frontier models (reasoning effort: low)}} \\
\quad GPT-6 Luna & \underline{85.1} & 14.9 & 0.0 & \underline{82.7} & 14.8 & 2.6 & 0.0 & \underline{92.2} & 7.8 & 0.0 \\
\quad GPT-6.1 Sol & \textbf{98.7} & 1.3 & 0.0 & \textbf{98.4} & 0.8 & 0.8 & 0.0 & \textbf{99.6} & 0.4 & 0.0 \\
\midrule
\multicolumn{11}{@{}l}{\textsc{Qwen3.5, four sizes (thinking off unless marked)}} \\
\quad Qwen3.5-2B & \textcolor{gray}{3.3} & 71.9 & 24.8 & 4.2 & 70.2 & 0.1 & 25.5 & 0.8 & 76.6 & 22.7 \\
\quad Qwen3.5-4B & \textcolor{gray}{14.1} & 84.9 & 1.0 & 8.3 & 73.8 & 16.5 & 1.3 & 30.9 & 69.1 & 0.0 \\
\quad Qwen3.5-9B & \textcolor{gray}{19.6} & 79.6 & 0.8 & 12.8 & 64.8 & 21.8 & 0.7 & 39.5 & 59.4 & 1.2 \\
\quad Qwen3.5-27B & 30.9 & 69.1 & 0.0 & 19.5 & 39.5 & 41.0 & 0.0 & 64.1 & 35.9 & 0.0 \\
\quad Qwen3.5-27B (think) & 80.3 & 13.9 & 5.8 & 76.9 & 14.5 & 2.0 & 6.6 & 90.2 & 6.2 & 3.5 \\
\midrule
\multicolumn{11}{@{}l}{\textsc{Bangla-adapted model and its base}} \\
\quad Gemma 3 12B & \textcolor{gray}{8.5} & 21.2 & 70.3 & 10.3 & 17.5 & 1.9 & 70.3 & 3.1 & 26.6 & 70.3 \\
\quad TigerLLM 12B & \textcolor{gray}{11.1} & 69.5 & 19.4 & 13.3 & 63.8 & 2.6 & 20.3 & 4.7 & 78.5 & 16.8 \\
\midrule
\multicolumn{11}{@{}l}{\textsc{Other open models}} \\
\quad Gemma 2 9B & \textcolor{gray}{9.7} & 86.8 & 3.5 & 12.2 & 84.0 & 0.8 & 3.0 & 2.3 & 92.6 & 5.1 \\
\quad Llama 3.1 8B & \textcolor{gray}{6.7} & 84.2 & 9.1 & 6.9 & 80.0 & 4.0 & 9.1 & 6.2 & 84.8 & 9.0 \\
\bottomrule
\end{tabular}
\label{tab:main}
\end{table*}

We discuss the experimental setup for the evaluation of the performance of different LLMs against \textsc{\name}. \mimon{NAME is not resolved to BanglaPLAN in the document.}\masud{fixed.}

\paragraph{Models.}
We evaluate 11 configurations across four groups, as shown in Table~\ref{tab:main}: two frontier models, GPT-6 Luna and GPT-6.1 Sol; Qwen3.5 at four sizes (2B, 4B, 9B, and 27B), with an additional thinking-enabled 27B configuration to assess test-time reasoning at fixed weights; the Bangla-adapted TigerLLM (12B)~\cite{raihan2025tigerllm} alongside Gemma~3 (12B); and two additional open models, Gemma~2 (9B) and Llama~3.1 (8B).
We use greedy decoding (temperature $=0$) for all configurations.

\paragraph{Protocol.}
We evaluate all models zero-shot using the released \textsc{\name} prompts, which specify the scenario, constraints, tie-breaking rule, and required JSON-array format. We set reasoning effort to \emph{low} for the two frontier models and disable thinking for all open models. We additionally evaluate Qwen3.5-27B with thinking enabled.
We discuss the prompt template in Appendix ~\ref{sec:eval-prompt}.

\paragraph{Scoring.}
An answer is \emph{correct} if it parses as the required JSON array and exactly matches the gold answer after canonicalization, as we discussed in Section~\ref{sec:benchmark}. The benchmark contains 744 feasible instances and 256 infeasible instances. Table~\ref{tab:main} reports overall accuracy and outcome breakdowns. \textbf{EM} (exact match), our primary metric, measures accuracy across all instances. For feasible instances, \textbf{Plan} measures correct-plan accuracy, \textbf{Wrong plan} measures schema-compliant but incorrect plans, and \textbf{Abstention} measures how often a model answers ``infeasible'' despite a valid plan existing. For infeasible instances, \textbf{Flag} measures correct ``infeasible'' responses, while \textbf{False plan} measures schema-compliant responses proposing a plan when none exists. \textbf{Format} measures outputs that violate the required JSON-array format or answer schema, reported overall and separately for feasible and infeasible instances.

EM combines plan construction and infeasibility detection:
\begin{equation}
\text{EM} = 0.744\cdot\text{Plan} + 0.256\cdot\text{Flag}.
\label{eq:decomp}
\end{equation}
Reporting these components separately shows whether a model succeeds by constructing valid plans or detecting infeasible requests. As a reference baseline, a policy that always returns ``infeasible'' in the required format achieves 25.6\% EM with no format errors. Models below this baseline perform worse than a fixed response that exploits the benchmark's label distribution.

\paragraph{English translations.}
To assess how prompt language affects performance, we translate all prompts from Bangla into English using Claude Sonnet 5.5. We preserve identifiers, numbers, times, and the JSON answer block verbatim, allowing the same gold answers to be used for both versions. We evaluate GPT-6.1 Sol and Qwen3.5-2B/4B/9B on both language versions.

\paragraph{Fine-tuning.}
To test whether lightweight supervision improves small-model performance, we fine-tune Qwen3.5-2B using LoRA (rank 16, $\alpha{=}32$, dropout 0.05) for 3 epochs, with a learning rate of $2{\times}10^{-4}$, 5\% warm-up, and a batch size of 4. The training target is the gold JSON answer, and loss is computed only on target tokens. We train on randomly selected 834 instances and report all fine-tuning results on the remaining 166 held-out instances, including 42 infeasible cases. We ran the fine-tuning experiment on a MacBook Pro with 64 GB of memory.

\section{Results}
\label{sec:results}
\masud{Fixed} \mimon{Thank you, vaiya. Maybe my overthinking mind is giving you some hard times; sorry for that.}
We address RQ1 through the overall results in Table~\ref{tab:main} and the score decomposition \S\ref{sec:res-overall}. We address RQ2 through breakdowns by difficulty, answer type, and failure mode, alongside the language and fine-tuning experiments (\S\ref{sec:res-variation}--\ref{sec:res-ft}).

\subsection{RQ1: Performance across models and reasoning settings}
\label{sec:res-overall}

\paragraph{Models with reasoning lead in exact match.} \mimon{This is great. An alternative could be "Reasoning models outperform the non-reasoning ones on \name."?}\masud{I liked your idea. Just thought that we may not say qwen3.5 models non-reasoning model. Since they are all reasoning model in general. But thinking off, they work as general model. So I rephrased the header.} \mimon{great point.}
Table~\ref{tab:main} shows a clear performance gap. GPT-6.1 Sol achieves the highest EM at \textbf{98.7\%}, followed by GPT-6 Luna at 85.1\% and Qwen3.5-27B with thinking at 80.3\%. These three configurations use test-time reasoning. In contrast, all other configurations score at or below 30.9\%, and \textbf{7 of 11} perform below the always-``infeasible'' baseline. \mimon{Should the line start with "In contrast, ..."?}
\masud{Fixed}

\begin{figure}[t]
\centering
\includegraphics[width=0.95\columnwidth]{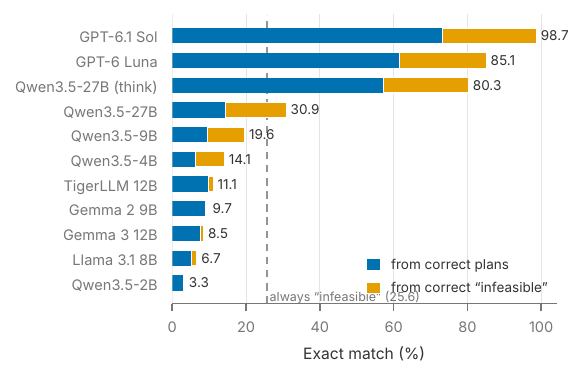}
\caption{Exact-match score decomposed into correct plans (blue) and ``correct infeasible'' answers (orange). The dashed line shows ``always-infeasible'' baseline.}
\label{fig:decomp}
\vspace{-10pt}
\end{figure}

\paragraph{Exact match masks different response behaviors.}
Similar EM scores can reflect different behaviors. \mimon{Not sure, but could we write it more clearly like, The capability of evaluated LLM models varies across task accomplishments of \name. Actually, I just wanted to ensure that a reviewer gets a clear idea (not vagueness \& technical detail) from the headline; then we can go into detail.}
Figure~\ref{fig:decomp} separates contributions from correct plans and correctly flagged infeasible instances. The three reasoning models earn most of their scores through correct plans. In contrast, Qwen3.5-27B without thinking earns \textbf{53\%} of its 30.9\% EM from correctly answering ``infeasible''. It returns this answer on 46.9\% of all instances, almost twice the benchmark's infeasible share of 25.6\%. Gemma, LLaMA, and TigerLLM show the opposite tendency: they rarely identify infeasible instances, with Flag scores no higher than 6.2\%. 
Appendix~\ref{sec:wrong-plan-audit} audits Wrong plan errors for two models, Qwen3.5-2B and GPT-6.1 Sol, and finds mostly hard-constraint violations for the former and more tie-break differences for the latter.

\paragraph{Thinking yields larger gains than scaling or Bangla adaptation.}
Within the Qwen3.5 family, increasing model size from 2B to 27B raises EM from 3.3\% to 30.9\%, averaging \textbf{7.4 percentage points per parameter doubling}. Plan accuracy improves more slowly, by 4.1 points per doubling. Enabling thinking with the \emph{same} 27B weights yields a larger gain than scaling across this range: EM rises by \textbf{49.4 points} to 80.3\%, Plan increases from 19.5\% to 76.9\%, and incorrect abstentions fall from 41.0\% to 2.0\%.

Bangla adaptation offers a much smaller gain. Compared with its base model, Gemma~3~12B, TigerLLM improves EM from 8.5\% to 11.1\%, alongside a substantial reduction in format errors from 70.3\% to 19.4\%. Both models remain below the always-``infeasible'' baseline.

\begin{takeaway}
\textbf{Observation 1.} Test-time reasoning outperforms scaling: reasoning configurations reach 80.3--98.7\% EM, versus at most 30.9\% for direct answers. Seven of eleven configurations fall below the 25.6\% always-``infeasible'' baseline.
\end{takeaway}

\begin{figure}[t]
\centering
\includegraphics[width=0.95\columnwidth]{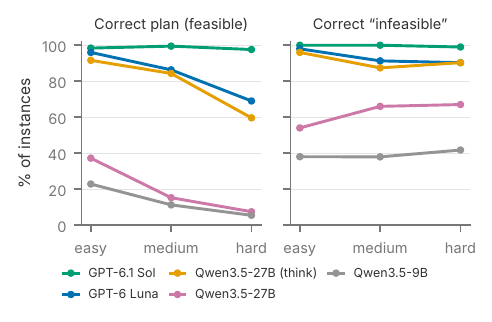}
\caption{Planning accuracy drops with difficulty, while infeasibility detection remains stable. Left: correct plans on feasible instances. Right: correct ``infeasible'' answers on infeasible instances.}
\label{fig:difficulty}
\vspace{-10pt}
\end{figure}

\subsection{RQ2: Performance across difficulty levels and failure modes}
\label{sec:res-variation}

\paragraph{Difficulty affects planning more than infeasibility detection.}
Figure~\ref{fig:difficulty} shows that plan accuracy declines sharply from easy to hard instances: GPT-6 Luna drops from 96.0\% to 69.0\%, and Qwen3.5-27B with thinking from 91.6\% to 59.6\%. Infeasibility detection remains more stable. GPT-6.1 Sol's Flag score falls from 100.0\% to 99.0\%, while Qwen3.5-27B with thinking declines from 96.0\% to 90.3\%. These results suggest that constructing a valid plan under more interacting constraints is harder than identifying an infeasible request. The gap is particularly clear for Qwen3.5-27B without thinking: Plan accuracy falls from 37.2\% to 7.4\%, while Flag rises from 54.0\% to 67.0\%. Better detection of infeasible cases therefore does not imply better plan construction.

\paragraph{Constructed answers are the hardest.} \mimon{Is it something like "LLM models often struggle in constructing structured/formatted answers."?}\masud{that is what I meant.}
Models perform better when choosing a label than when constructing a structured answer (Appendix Table~\ref{tab:app-kind}). Qwen3.5-27B with thinking achieves 99.0\% accuracy on classification but only 68.6\% on schedules, its weakest answer type. Models without thinking show the same pattern at lower accuracy levels. Results by task family are consistent (Appendix Table~\ref{tab:app-family}): single-label eligibility decisions (F01) are easiest for every model except GPT-6.1 Sol, which performs near the ceiling. Procurement (F08), sequencing (F06), and appointments (F03) are among the hardest families, requiring models to select items within a budget, respect dependencies, or coordinate timed bookings. In contrast, language register has little effect (Appendix Table~\ref{tab:app-register}): GPT-6 Luna ranges from 82.5\% (code-mixed) to 87.0\% (formal), and Qwen3.5-27B with thinking stays between 79.0\% and 81.0\%. Difficulty comes from what the answer requires, not the presentation.

\jami{I may have missed, but did we specify what format error we are talking about} 
\masud{we mentioned about Format error in Section 4 Evaluation-> Scoring}
\paragraph{Format errors
limit performance.}
Gemma~3 fails to follow the required format on \textbf{70.3\%} of instances. Even among correctly formatted outputs, its accuracy is only 28.6\%. For comparison, Qwen3.5-27B, the strongest open model without thinking, achieves 30.9\% accuracy across all instances. Qwen3.5-2B produces format errors on 24.8\% of instances, rising from 14.0\% on easy problems to 32.0\% on hard problems, where answers are longer.

\begin{takeaway}
\textbf{Observation 2.} Hard instances break plan construction, not infeasibility detection. Answer kinds that must be constructed (schedules, selections, ordered steps) are the hardest.
\end{takeaway}

\begin{table}[t]
\centering\small
\setlength{\tabcolsep}{4.5pt}
\caption{Bangla prompts vs.\ their English translations (same instances, same scoring) English improves Qwen EM but yields limited gains in plan accuracy; GPT-6.1 Sol performs better in Bangla.\jami{Normally, the caption stays at the top for tables, right? And if a reviewer comes directly to this table, what's the takeaway that English translation improves the performance; thus do we advocate translation}\masud{fixed}
}
\begin{tabular}{@{}l rr rr rr@{}}
\toprule
 & \multicolumn{2}{c}{EM$\uparrow$} & \multicolumn{2}{c}{Plan$\uparrow$} & \multicolumn{2}{c}{Flag$\uparrow$} \\
\cmidrule(lr){2-3}\cmidrule(lr){4-5}\cmidrule(l){6-7}
Model & bn & en & bn & en & bn & en \\
\midrule
GPT-6.1 Sol & \textbf{98.7} & 97.4 & \textbf{98.4} & 96.6 & 99.6 & 99.6 \\
Qwen3.5-2B & 3.3 & \textbf{5.8} & 4.2 & \textbf{6.6} & 0.8 & \textbf{3.5} \\
Qwen3.5-4B & 14.1 & \textbf{22.4} & 8.3 & \textbf{8.6} & 30.9 & \textbf{62.5} \\
Qwen3.5-9B & 19.6 & \textbf{24.3} & 12.8 & \textbf{15.2} & 39.5 & \textbf{50.8} \\
\bottomrule
\end{tabular}
\label{tab:language}
\vspace{-12pt}
\end{table}

\subsection{Bangla vs.\ English prompts}
\label{sec:res-lang}
\paragraph{Bangla -> English translation does not consistently improve performance.}
Table~\ref{tab:language} shows that English translation does not consistently improve performance. GPT-6.1 Sol performs slightly better in Bangla, achieving 98.7\% EM compared with 97.4\% in English. The smaller Qwen models gain 2.5, 8.3, and 4.7 EM points at 2B, 4B, and 9B, respectively, but the score decomposition shows limited improvement in plan construction.

\paragraph{Language gains mainly reflect formatting and infeasibility responses.}
Qwen3.5-4B illustrates this distinction. Plan accuracy changes little, from 8.3\% in Bangla to 8.6\% in English, while ``infeasible'' responses become much more frequent: Flag rises from 30.9\% to 62.5\%, and incorrect Abstention rises from 16.5\% to 48.3\%. English therefore improves infeasibility detection at the cost of rejecting more feasible requests. For Qwen3.5-2B, English roughly halves format errors, from 24.8\% to 12.0\%. Overall, language gains mainly reflect better formatting and more infeasibility responses; small models still struggle to construct plans in both languages.

\subsection{Fine-tuning improves format but offers little planning gain}
\label{sec:res-ft}

Fine-tuning Qwen3.5-2B on gold answers substantially improves overall accuracy. Table~\ref{tab:finetune} shows: on the 166 held-out instances, EM rises from 2.4\% to \textbf{27.1\%}, format errors fall from 18.7\% to zero, and Flag increases from zero to 95.2\%. However, Plan accuracy improves only from 3.2\% to 4.0\%. The model answers ``infeasible'' on 90.9\% of instances, although only 25.3\% are infeasible, exceeding the always-``infeasible'' baseline by just 1.8 points.
This response pattern may reflect the training targets: infeasible instances share one short answer, whereas feasible instances require longer, distinct plans.

\begin{table}[t]
\centering\scriptsize
\setlength{\tabcolsep}{3pt}
\caption{Fine-tuning Qwen3.5-2B on gold answers (\%), scored on the 166 held-out instances,
42 of them infeasible, so always answering ``infeasible'' scores 25.3.}
\begin{tabular}{@{}l r rr r r@{}}
\toprule
 & & \multicolumn{2}{c}{Feasible} & Infeasible & \\
\cmidrule(lr){3-4}\cmidrule(lr){5-5}
 & EM$\uparrow$ & Plan$\uparrow$ & Abstention & Flag$\uparrow$ & Format \\
\midrule
\textit{Always ``infeasible''} & \textit{25.3} & \textit{0.0} & \textit{100.0} & \textit{100.0} & \textit{0.0} \\
Qwen3.5-2B, untrained & 2.4 & 3.2 & 0.0 & 0.0 & 18.7 \\
\;\; + SFT (gold answers) & 27.1 & 4.0 & 89.5 & 95.2 & 0.0 \\
\bottomrule
\end{tabular}
\label{tab:finetune}
\vspace{-10pt}
\end{table}

\section{Conclusion}
\label{sec:conclusion}
We introduced \textsc{\name} for evaluating constraint-based planning in Bangla with validated answers and deterministic scoring. GPT-6.1 Sol achieves 98.7\% accuracy, while smaller models struggle with planning and formatting. Test-time reasoning raises Qwen3.5-27B's accuracy from 30.9\% to 80.3\%. English prompts, Bangla adaptation, and gold-answer fine-tuning reduce format errors but yield limited planning gains. Future work should explore reasoning-focused training, including reinforcement learning with rewards for verified plans, to improve small open models' reliability in Bangla.

\section*{Limitations}
\textsc{\name} is grounded in Bangladesh and written in its registers of Bangla; its institutions and conventions may not transfer to other Bangla-speaking regions.
Each problem is pinned to one canonical answer by stating objectives and tie-breaking rules explicitly. This makes scoring deterministic, but it removes the open-endedness of real requests, and exact match gives no credit to nearly correct plans.
Our difficulty labels count binding constraints and supporting facts; we do not validate them against human solving effort.
Claude assisted scenario drafting and the final consistency check, so the problems may carry some of its stylistic tendencies, although humans wrote, validated, and adjudicated every item.
All models are evaluated zero-shot with one prompt; few-shot examples, other prompt formats, or higher reasoning effort could change the numbers.
Finally, our fine-tuning experiment uses one 2B model and one recipe, so the negative result on answer-only supervision is suggestive rather than general.

\bibliography{bib/papers}


\appendix
\newpage

\section{Prompt Template}
\label{sec:eval-prompt}

Each model receives, for every instance, the following prompt as a
single user turn:

\begin{figure}[h]
\centering
\begin{tikzpicture}
\node[draw=promptgreen, dashed, line width=0.9pt, rounded corners=8pt,
      inner sep=10pt, text width=0.93\columnwidth, align=left,
      font=\ttfamily\small] {%
\ph{gold\_prompt}%
};
\end{tikzpicture}
\caption{The evaluation prompt: the instance's \texttt{gold\_prompt},
sent verbatim.}
\label{fig:eval-prompt}
\end{figure}

\noindent
That is, the instance's \texttt{gold\_prompt} is the entire input. We
add no system prompt, no in-context exemplars, and no chain-of-thought
instruction; API models receive it as the sole user message, and
open-weight models receive it through their own chat template. The
prompt itself is self-contained: it states the scenario, the facts and
constraints, an objective with a deterministic tie-break rule where
needed, the JSON-array output schema, and the infeasibility marker
\texttt{[\{"is\_feasible": false\}]}, so each instance has exactly one
correct answer.

\section{Investigating \emph{Wrong plan} Errors}
\label{sec:wrong-plan-audit}

Exact match labels any parsed plan on a feasible instance that differs from the gold plan as \emph{Wrong plan}. This category may contain three kinds of error: a hard-constraint violation (V), a feasible plan that loses on the stated objective (S), or an optimal plan that differs only on a tie-break or output convention (N). Because the constraints are written in Bangla prose, we manually audited errors from a low-scoring (Qwen3.5-2B)  and a high-scoring model (GPT-6.1 Sol). Table~\ref{tab:wrong-plan-examples} presents ten such audited cases.

\paragraph{Qwen3.5-2B.}
Qwen3.5-2B produced 522 wrong plans across the 1,000 instances. We randomly sampled 50, stratified in proportion to answer type: 15 schedules, 8 routes, 7 ordered-step plans, 6 classifications, 5 selections, 4 allocations, 3 matchings, and 2 rankings. An author checked each model output against the prompt's hard constraints, objectives, and tie-breaking rules. 

The 50 sampled Qwen3.5-2B plans comprise 49~V, 0~S, and 1~N. Common violations include copying \texttt{2026-11-11} from the output example instead of using the requested date, scheduling before an office opens, and ignoring stated durations or prerequisites. Some exceed limits substantially: one route costs 3,700~BDT against a 2,100~BDT budget; another plan seats 3,300 candidates from a pool of 3,000. Other errors include selecting excluded options, assigning the same slot twice, and inventing identifiers. The sole N case is a four-patient triage ranking that satisfies the constraints but swaps the final two patients against the ledger-order tie-break. In this sample, \emph{Wrong plan} therefore almost always indicates a constraint violation rather than a near-canonical answer.

\paragraph{GPT-6.1 Sol.}
GPT-6.1 Sol misses 13 of 1,000 instances, and here the
composition is different. Six are wrong plans: three meet the constraints and objective but fail a tie-break, one is feasible but suboptimal, and two violate a hard constraint. In two cases, the model schedules an SMS check on a later working day despite an earlier one being available. In the third, it starts a supply slot at 11:20 although the line is free from 10:00. In the suboptimal plan, an extra sack takes the costlier river route, raising transport cost from 6,925 to 6,960~BDT. In one case, the model starts an ultrasound 40 minutes after a blood test, although the prompt requires a 45-minute wait. In another, the team returns at 20:50, missing the 20:30 deadline. A further miss proposes a plan for an infeasible instance, exceeding its 61,799~BDT budget by one taka while reporting a total of 60,300~BDT.

\begin{table*}[t]
\centering
\small
\setlength{\tabcolsep}{4pt}
\renewcommand{\arraystretch}{1.15}
\caption{Ten audited \emph{Wrong plan} cases. V: hard-constraint violation; S: feasible but worse on the objective; N: optimal but fails a tie-break. Instance IDs omit the \texttt{BP-} prefix. The audit sheets include the relevant rules in Bangla.}
\begin{tabular}{@{}l l l p{5.2cm} p{5.6cm} c@{}}
\toprule
\textbf{Model} & \textbf{Instance} & \textbf{Kind} & \textbf{Rule in the prompt} & \textbf{What the model's plan does} & \textbf{Verdict} \\
\midrule
Qwen3.5-2B & F04-T038-001 & route & Fare at most 2,100~BDT; patient needs oxygen, which the microbus lacks; arrival by 02:00 & Takes ambulance and microbus for 3,700~BDT; arrives at 02:05 & V \\
\midrule
Qwen3.5-2B & F03-T025-007 & schedule & Plan for 9--11 December; 60 minutes between villages; no overlaps & Dates every action 2026-11-11, copied from the format example; allows no travel time & V \\
\midrule
Qwen3.5-2B & F02-T015-005 & ordered steps & Hospital slip before application; verification at least 24~h later, on Sun/Tue & Applies before obtaining the slip; verifies immediately on a Monday & V \\
\midrule
Qwen3.5-2B & F01-T002-008 & classification & International track requires English score $\geq 6.5$; applicant has 6.0 & Selects the international track & V \\
\midrule
Qwen3.5-2B & F08-T078-007 & selection & One bidder per lot; bidder~1 fails the copier test; bidder~3 is blacklisted; capacity is 40 pieces & Selects all three bidders, including ineligible copiers, and buys 75 desk sets and 24 cabinets & V \\
\midrule
Qwen3.5-2B & F10-T097-002 & matching & Exactly one unit per slot; list each slot once & Assigns two engines to the same slot and lists it twice & V \\
\midrule
Qwen3.5-2B & F10-T094-001 & ranking & Chest pain before crash; crash before breathing; at most one wound patient in the first call; ledger-order tie-break & Satisfies the constraints but swaps the final two patients relative to ledger order & N \\
\midrule
\midrule
GPT-6.1 Sol & F03-T022-008 & schedule & Ultrasound at least 45 minutes after blood test ends at 08:20 & Starts ultrasound at 09:00, after only 40 minutes & V \\
\midrule
GPT-6.1 Sol & F07-T070-008 & allocation & Minimise transport cost; road costs 20~BDT/sack, river 35; satisfy minimums and stock limits & Moves one extra sack by river to honor a preference; cost rises from 6,925 to 6,960~BDT & S \\
\midrule
GPT-6.1 Sol & F03-T026-009 & schedule & SMS check at least two days after biometrics, within 13--20 Oct, on a working day; earliest-start tie-break & Checks on 15 Oct instead of the earliest feasible date, 13 Oct & N \\
\bottomrule
\end{tabular}
\label{tab:wrong-plan-examples}
\end{table*}

\section{Annotator and Validator Guidelines}
\label{sec:annotation-guidelines}

\paragraph{Guideline for Annotators.}
\begin{itemize}[leftmargin=15pt, nosep]
    \item Write prompts directly in natural Bangla using the assigned task family, register, and answer type. Identify scenario rules as fictional.
    \item State all entities, quantities, constraints, objectives, and tie-breaking rules explicitly. Provide reference dates and calendar-conversion information where needed, and distinguish optional preferences from mandatory constraints.
    \item For infeasible variants, introduce a binding conflict and verify that no valid plan remains. Claim minimal infeasibility only when relaxing the designated constraint restores feasibility.
    \item Derive the gold answer using the stated constraints, ordered objectives, and tie-breaking rules. Follow the required JSON schema, identifiers, digits, timestamps, and ordering conventions; omit zero-valued allocations.
    \item Record the binding constraints and supporting facts needed to solve the problem. Assign difficulty by their total count $k$: easy ($k\leq10$), medium ($10<k\leq20$), or hard ($k>20$).
\end{itemize}

\paragraph{Guideline for Validators.}
\begin{itemize}[leftmargin=15pt, nosep]
    \item Solve the Bangla prompt independently without viewing its gold answer. Check clarity, completeness, feasibility, objectives, tie-breaking, and schema compliance.
    \item Compare the independently derived answer with the gold answer. For infeasible items, verify that the constraints rule out every valid plan.
    \item Resolve disagreements with the author and validate revised items again. Treat model-generated mismatches and suggestions as review signals; humans determine the final prompt and gold answer.
\end{itemize}

\section{Results by Task Family and Answer Kind}
\label{sec:appendix-breakdowns}

Tables~\ref{tab:app-family}--\ref{tab:app-register} report exact match on Bangla instances for all eleven configurations, broken down by task family, answer kind, and language register, respectively.

\begin{table*}[t]
\centering\footnotesize
\setlength{\tabcolsep}{3.5pt}
\caption{Exact match (\%) by task family (bn). \textbf{Bold} = best, \underline{underline} = second best per row.
Procurement (F08), sequencing (F06), and appointments (F03) are hardest, requiring models to select items within a budget, respect dependencies, or coordinate timed bookings.\jami{Should we also briefly add the take way of the message in text and do you think caption should be at top}\masud{fixed}}
\begin{tabular}{@{}lrrrrrrrrrrrr@{}}
\toprule
 & $n$ & \rot{GPT-6 Luna} & \rot{GPT-6.1 Sol} & \rot{Qwen3.5-2B} & \rot{Qwen3.5-4B} & \rot{Qwen3.5-9B} & \rot{Qwen3.5-27B} & \rot{Qwen3.5-27B (think)} & \rot{Gemma 3 12B} & \rot{TigerLLM 12B} & \rot{Gemma 2 9B} & \rot{Llama 3.1 8B} \\
\midrule
F01 Eligibility & 100 & \underline{98.0} & 97.0 & 21.0 & 34.0 & 46.0 & 64.0 & \textbf{99.0} & 43.0 & 45.0 & 40.0 & 34.0 \\
F02 Procedures & 100 & \underline{88.0} & \textbf{100.0} & 0.0 & 7.0 & 10.0 & 24.0 & 77.0 & 0.0 & 0.0 & 2.0 & 2.0 \\
F03 Appointments & 100 & \underline{78.0} & \textbf{97.0} & 2.0 & 15.0 & 19.0 & 28.0 & 68.0 & 0.0 & 5.0 & 6.0 & 4.0 \\
F04 Journeys & 100 & \underline{94.0} & \textbf{100.0} & 1.0 & 19.0 & 27.0 & 30.0 & 88.0 & 0.0 & 9.0 & 2.0 & 4.0 \\
F05 Rosters & 100 & \underline{80.0} & \textbf{100.0} & 0.0 & 17.0 & 19.0 & 30.0 & 77.0 & 4.0 & 9.0 & 7.0 & 4.0 \\
F06 Sequencing & 100 & \underline{81.0} & \textbf{99.0} & 0.0 & 10.0 & 13.0 & 23.0 & 68.0 & 0.0 & 2.0 & 4.0 & 0.0 \\
F07 Allocation & 100 & \underline{88.0} & \textbf{99.0} & 2.0 & 11.0 & 16.0 & 29.0 & 87.0 & 15.0 & 14.0 & 16.0 & 8.0 \\
F08 Procurement & 100 & 78.0 & \textbf{99.0} & 0.0 & 3.0 & 9.0 & 10.0 & \underline{83.0} & 0.0 & 1.0 & 0.0 & 0.0 \\
F09 Logistics & 100 & \underline{79.0} & \textbf{98.0} & 0.0 & 5.0 & 11.0 & 30.0 & 76.0 & 2.0 & 3.0 & 1.0 & 3.0 \\
F10 Emergency & 100 & \underline{87.0} & \textbf{98.0} & 7.0 & 20.0 & 26.0 & 41.0 & 80.0 & 21.0 & 23.0 & 19.0 & 8.0 \\
\bottomrule
\end{tabular}
\label{tab:app-family}
\end{table*}

\begin{table*}[t]
\centering\footnotesize
\setlength{\tabcolsep}{3.5pt}
\caption{Exact match (\%) by answer kind (bn). \textbf{Bold} = best, \underline{underline} = second best per row.
GPT-6.1 Sol leads on seven of eight answer kinds, while all other models achieve their highest exact match on classification.
\jami{Should we also briefly add the take way of the message in text and do you think the caption should be at top}\masud{fixed}
}
\begin{tabular}{@{}lrrrrrrrrrrrr@{}}
\toprule
 & $n$ & \rot{GPT-6 Luna} & \rot{GPT-6.1 Sol} & \rot{Qwen3.5-2B} & \rot{Qwen3.5-4B} & \rot{Qwen3.5-9B} & \rot{Qwen3.5-27B} & \rot{Qwen3.5-27B (think)} & \rot{Gemma 3 12B} & \rot{TigerLLM 12B} & \rot{Gemma 2 9B} & \rot{Llama 3.1 8B} \\
\midrule
allocation & 160 & \underline{88.8} & \textbf{99.4} & 1.2 & 10.0 & 13.8 & 28.7 & 84.4 & 8.1 & 10.0 & 10.6 & 1.2 \\
classification & 100 & \underline{98.0} & 97.0 & 21.0 & 34.0 & 46.0 & 64.0 & \textbf{99.0} & 43.0 & 45.0 & 40.0 & 34.0 \\
matching & 60 & \underline{95.0} & \textbf{100.0} & 5.0 & 25.0 & 26.7 & 43.3 & 93.3 & 30.0 & 30.0 & 28.3 & 20.0 \\
ordered steps & 140 & \underline{88.6} & \textbf{100.0} & 0.0 & 6.4 & 8.6 & 23.6 & 74.3 & 0.0 & 0.0 & 1.4 & 1.4 \\
ranking & 40 & \underline{80.0} & \textbf{95.0} & 10.0 & 22.5 & 32.5 & 40.0 & 72.5 & 27.5 & 20.0 & 12.5 & 10.0 \\
route & 100 & \underline{94.0} & \textbf{100.0} & 1.0 & 19.0 & 27.0 & 30.0 & 88.0 & 0.0 & 9.0 & 2.0 & 4.0 \\
schedule & 290 & \underline{76.2} & \textbf{97.9} & 0.7 & 12.1 & 17.2 & 28.6 & 68.6 & 0.0 & 4.8 & 4.8 & 3.1 \\
selection & 110 & 75.5 & \textbf{99.1} & 0.0 & 3.6 & 9.1 & 10.0 & \underline{84.5} & 0.0 & 0.9 & 0.0 & 0.0 \\
\bottomrule
\end{tabular}
\label{tab:app-kind}
\end{table*}


\begin{table*}[t]
\centering\footnotesize
\setlength{\tabcolsep}{3.5pt}
\caption{Exact match (\%) by language register (bn). \textbf{Bold} = best, \underline{underline} = second best per row.
GPT-6.1 Sol leads on every register. Among the three strongest models, GPT-6 Luna varies most, scoring lowest on code-mixed prompts (82.5\%)}
\begin{tabular}{@{}lrrrrrrrrrrrr@{}}
\toprule
 & $n$ & \rot{GPT-6 Luna} & \rot{GPT-6.1 Sol} & \rot{Qwen3.5-2B} & \rot{Qwen3.5-4B} & \rot{Qwen3.5-9B} & \rot{Qwen3.5-27B} & \rot{Qwen3.5-27B (think)} & \rot{Gemma 3 12B} & \rot{TigerLLM 12B} & \rot{Gemma 2 9B} & \rot{Llama 3.1 8B} \\
\midrule
code-mixed & 200 & \underline{82.5} & \textbf{99.0}  & 4.0 & 17.0 & 21.0 & 29.0 & 79.0 & 9.0 & 12.0 & 11.0 & 8.5 \\
colloquial & 100 & \underline{87.0} & \textbf{100.0} & 1.0 & 18.0 & 25.0 & 29.0 & 81.0 & 7.0 & 8.0  & 9.0  & 6.0 \\
everyday   & 500 & \underline{85.0} & \textbf{98.8}  & 4.0 & 12.4 & 18.8 & 31.0 & 80.8 & 9.8 & 12.4 & 9.8  & 6.2 \\
formal     & 200 & \underline{87.0} & \textbf{97.5}  & 2.0 & 13.5 & 17.5 & 33.5 & 80.0 & 5.5 & 8.5  & 8.5  & 6.5 \\
\bottomrule
\end{tabular}
\label{tab:app-register}
\end{table*}

\section{Acknowledgements}
\paragraph{AI Assistance Disclosure.}
We used generative AI tools, including ChatGPT and Claude, for language editing and paraphrasing of author-written text. Claude also suggested scenario details during benchmark drafting and flagged potential inconsistencies for human review. Human annotators selected and revised the scenarios, independently validated the problems, and determined the final gold answers. All content was reviewed and revised by the authors, who take full responsibility for the paper's correctness and citations.

\end{document}